%% file: root.tex
\documentclass[letterpaper, 10 pt, conference]{ieeeconf}  

\usepackage{amssymb}
\usepackage{amsmath}
\usepackage{commath}
\usepackage{mathtools}

\usepackage{hyperref}
\usepackage{import}
\usepackage{paralist}
\usepackage{svg}
\usepackage{textcomp}
\usepackage{gensymb}
\usepackage{dsfont}
\usepackage{subfig}

\IEEEoverridecommandlockouts                              

\usepackage{graphics} 
\usepackage{epsfig} 
\usepackage{todonotes}
\usepackage[normalem]{ulem}

\title{\LARGE \bf
Whole-Body Walking Generation using Contact Parametrization:\\
A Non-Linear Trajectory Optimization Approach
}

\author{Stefano Dafarra${}^{1,2}$, Giulio Romualdi${}^{1,2}$, Giorgio Metta${}^{1}$, and Daniele Pucci${}^{1}$
\thanks{This project has received funding from the European Union’s Horizon	2020 research and innovation programme under grant agreement No. 731540	(An.Dy).}%
\thanks{${}^{1}$ Dynamic Interaction Control, Istituto Italiano di Tecnologia, 16163 Genova,
Italy (e-mail: name.surname@iit.it)}%
\thanks{${}^{2}$ Universit\`a degli Studi di Genova, DIBRIS, Genova, Italy}%
}

\DeclareMathOperator*{\minimize}{minimize}

\begin{document}
	
\maketitle
\thispagestyle{empty}
\pagestyle{empty}

\begin{abstract}
In this paper, we describe a planner capable of generating walking trajectories by using the centroidal dynamics and the full kinematics of a humanoid robot model. The interaction between the robot and the walking surface is modeled explicitly through a novel contact parametrization. The approach is \emph{complementarity-free} and does not need a predefined contact sequence. 
By solving an optimal control problem we obtain walking trajectories. In particular, through a set of constraints and dynamic equations, we model the robot in contact with the ground. We describe the objective the robot needs to achieve with a set of tasks. 
The whole optimal control problem is transcribed into an optimization problem via a Direct Multiple Shooting approach and solved with an off-the-shelf solver.
We show that it is possible to achieve walking motions automatically by specifying a minimal set of references, such as a constant desired Center of Mass velocity and a reference point on the ground.
\end{abstract}

\import{tex/}{Introduction}
\import{tex/}{SystemModeling}
\import{tex/}{Tasks}
\import{tex/}{OptimalControlProblem}
\import{tex/}{Experiments}

\import{tex/}{Conclusions}

\addtolength{\textheight}{0cm}   

\addcontentsline{toc}{section}{References}

\bibliography{IEEEabrv,Bibliography}

\end{document}

%% file: tex/Introduction.tex
\section{Introduction}
Planning locomotion trajectories for humanoid robots requires considering high-dimensional multi-body systems instantiating contacts with the surrounding environment. Given their intrinsic under-actuation, these robots have to exploit the interaction with the environment and their ability to change ``shape'' in order to move. 

During the DARPA Robotics Challenge~\cite{feng2015optimization}, it became popular to approach the locomotion problem with a hierarchical control architecture. In our previous works \cite{dafarra2018control, romualdi2018benchmarking} we explored this concept by adopting a three layer control architecture. The outer layer, i.e. the \emph{trajectory optimization for foot-step planning}, is in charge of generating walking trajectories starting from high-level commands, such as those coming from a joystick. The output of this layer is served to the \emph{receding horizon controller (RHC)}, also referred to as Model Predictive Control (MPC)~\cite{Mayne2000Stability}. Its aim is to generate \emph{centroidal} \cite{orin08} quantities obtainable by the robot without incurring into an undesired fall state. Compared to the first, this second layer adopts more complex models with a shorter prediction horizon. The last stage, the \emph{whole-body quadratic programming control} is in charge of stabilizing the planned trajectories exploiting the full robot model with a suitable \emph{Quadratic Programming} formulation. 

In this paper, we merge the first two layers, generating locomotion trajectories adopting the full kinematics of the robot and the centroidal dynamics. The approach follows what presented by authors of \cite{dai2014whole}. Nevertheless, no prior knowledge is injected on the system to generate walking trajectories, but the whole-body motions result from a particular choice of cost function. 

 When planning locomotion trajectories, the definition of contacts plays a central role. Several strategies are available in literature, here roughly summarized in four categories.

\emph{Fixed contact sequence, timing and location}. A common approach consists of assuming to know in advance where the contacts will be established and in which instant \cite{fernbach2018croc,herzog2015trajectory,posa2016optimization, de2010feature,dafarra2017receding}. Such choice simplifies the planning problem, leading to a lower computational effort. However, they need to rely on external contact planners. 

\emph{Predefined contact sequence}. During locomotion, it can be assumed to know in advance the contact sequence. As an example, for a biped robot, it can be assumed that a contact with the left foot will be followed by another one with the right foot. In other words, the phases are predefined while the remaining quantities (like positions and timings) are an output of the planner \cite{winkler18,carpentier2016versatile,caron2017make}. By specifying a different set of equations depending on the contact state, the hybrid nature arising from the establishment of contacts is easily modeled. The time spent by each phase can be turned into an optimization variable. Nevertheless, in case several point contacts, the definition of the various phases could become intractable. 

\emph{Mixed Integer Programming}. Instead of receiving the contact sequence as input, it is possible to use integer variables to determine when a particular contact has to be considered active or not \cite{deits2014footstep,aceituno2018simultaneous}. This approach requires \emph{Mixed Integer Programming} tools. While providing enhanced modeling capabilities, the exploitation of integer variables strongly affects the computational performances, especially in case several contacts are available. In addition, the availability of specialized solvers is limited.

\emph{Complementarity-free}. Authors of \cite{todorov2011convex, drumwright2010modeling} presented an approach which allows simulating multi-body systems subject to contacts, without enforcing complementarity conditions directly. Equivalently accurate results are obtained by maximizing the rate of energy dissipation. Such approach can be used to generate complex movements \cite{mordatch2012discovery, tassa2012synthesis}.

In this paper, we present a planner where neither contact sequences, locations or timings are fixed a priori. Additionally, we adopt a complementarity-free approach. Through a tailored parametrization of contacts, we impose complementarity conditions indirectly. Since the full robot kinematics is used, planned footsteps are within the robot work-space.

%% file: tex/SystemModeling.tex
\subsection{Notation}
\begin{itemize}
	\item The $i_{th}$ component of a vector $x$ is denoted as $x_i$.	
    \item The transpose operator is denoted by $(\cdot)^{\top}$.
	\item $\mathcal{I}$ is a fixed inertial frame with respect to (w.r.t.) 
	which the robot's absolute pose is measured. Its $z$ axis points against gravity, while the $x$ direction points forward.
	\item Given a function of time $f(t)$ the dot notation denotes the time derivative, i.e.
	$\dot{f} := \frac{\dif f}{\dif t}$. Higher order derivatives are denoted with a corresponding amount of dots.
	\item $\mathds{1}_n \in \mathbb{R}^{n \times n}$ denotes the identity matrix of dimension $n$.
	\item $0_{n \times n} \in \mathbb{R}^{n \times n}$ denotes a zero matrix.
	\item $S(\cdot)$ is the skew-symmetric operation associated with the cross product in $\mathbb{R}^3$.
	\item The weighted L2-norm of a vector $v \in \mathbb{R}^n$ is denoted by $\|v\|_W$, where $W \in \mathbb{R}^{n\times n}$ is a weight matrix.
	\item $^{A}R_{B} \in SO(3)$ and $^{A}H_{B} \in SE(3)$ denote the rotation and transformation matrices which transform a vector expressed in the $B$ frame into one expressed in $A$.
	\item $\mathbf{n}(\cdot), \mathbb{R}^3 \rightarrow \mathbb{R}^3$ returns the direction normal to the walking plane given the argument's $x$ and $y$ coordinates.
	\item $\mathbf{t}(\cdot), \mathbb{R}^3 \rightarrow \mathbb{R}^{3\times2}$ returns two perpendicular directions normal to $\mathbf{n}(\cdot)$. The composition of $\mathbf{t}(\cdot)$ and $\mathbf{n}(\cdot)$, $\left[\mathbf{t}(\cdot)~\mathbf{n}(\cdot)\right]$, defines the rotation matrix ${}^\mathcal{I}R_{plane}$.
	\item The function $h(p), \mathbb{R}^3 \rightarrow \mathbb{R}$ defines the distance between $p$ and the walking surface. 
	\item $e_1 := [1,0,0]^\top$, $e_2 := [0,1,0]^\top$ and $e_3 := [0,0,1]^\top$ denote the canonical basis vectors of $\mathbb{R}^3$.
	\item ${}^DV_{A,D} \in \mathbb{R}^6$ is the relative velocity between frame $A$ and $D$,  whose coordinates are defined in frame $D$.
	\item $diag(\cdot), \mathbb{R}^n \rightarrow \mathbb{R}^{n \times n}$ is a function casting the argument into the corresponding diagonal function.
\end{itemize}

\section{System Modeling} \label{sec:modeling}

\subsection{Contact Interface Modeling} \label{sec:contacts_interface}

When performing a step, the foot can impact the ground in a not flat configuration, reducing the amount of contact wrenches obtainable from the ground. At the same time, toe-off motions can be used to increase the work-space available during double support phases \cite{griffin2018straight}. Given these reasons, all the various contact configurations should be taken into account when planning step motions.

In order to reduce the complexity, it is possible to consider the foot as composed by a set of points, for example four points located at the corners of the foot \cite{dai2014whole, caron2017make}. Thanks to this choice, the several contact configurations can be modeled independently, depending on the number of points in contact.

A pure force can be applied on each contact point. In case of four points, twelve variables define a six dimensional quantity, i.e. the resulting contact wrench acting on the foot. This is a drawback that will be addressed later in Sec. \ref{sec:forceRegularization}.

\subsection{Contacts Force Constraints}
Define ${}_ip \in \mathbb{R}^3$ as the $i-$th contact point location in an inertial frame $\mathcal{I}$, and ${}_if \in \mathbb{R}^3$ as the force exerted on that point. Such force is expressed on a frame located in ${}_ip$ and with orientation parallel to $\mathcal{I}$. Since it results from the interaction of the foot with the ground, it is subject to constraints. Being a reaction force, its normal component with respect to the walking ground is supposed to be non-negative. In particular, $\mathbf{n}({}_ip)^\top{}_if \geq 0$.
Additionally, in order to avoid slippage, friction constraints should be satisfied:
\begin{equation} \label{eq:friction}
	\|\mathbf{t}({}_ip)^\top {}_if \| \leq \rho ~\mathbf{n}({}_ip)^\top{}_if
\end{equation}
where $\rho$ is the static friction coefficient.

\subsection{Contact Parametrization} \label{sec:contact_parametrization}

The contact force ${}_if$ applied to the $i-$th contact point is supposed to be not-null only if the point is in contact with the walking surface. This condition could be represented by the following equality:
\begin{equation}\label{eq:complementarity}
	h({}_i p) ~\mathbf{n}({}_ip)^\top{}_if = 0.
\end{equation}
Such constraint can be difficult to be tackled in an optimization framework. This is due to the fact that the feasible set is only constituted by two lines, namely $h({}_i p) = 0$ and $\mathbf{n}({}_ip)^\top{}_if = 0$, which are intersecting in the origin. In this point, the constraint Jacobian gets singular, thus violating the \emph{linear independence constraint qualification} (LICQ), on which most off-the-shelf solvers rely upon \cite{betts2010practical}.

In order to avoid the complications related to Eq. \eqref{eq:complementarity}, we adopt a simple parametrization. In particular, we assume to have full control over the derivative of both contact point positions and forces:
\begin{IEEEeqnarray}{RCL}
\phantomsection \IEEEyesnumber
{}_i\dot{p} &=& u_{{}_ip} \IEEEyessubnumber \label{eq:velocity_control}\\
{}_i\dot{f} &=& u_{{}_if}, \IEEEyessubnumber
\end{IEEEeqnarray}
where $u_{{}_ip}, ~ u_{{}_if} \in \mathbb{R}^3$ are control inputs. Then, we can impose Eq. \eqref{eq:complementarity} dynamically through the following constraints:
\begin{IEEEeqnarray}{RCLR}
	\phantomsection \IEEEyesnumber \label{eq:force_control_cases}
	-M_f \leq &u_{{}_if} & \leq M_f     & \quad \text{if } h({}_i p) = 0, \IEEEyessubnumber \label{eq:force_control_bounds}\\
			  &u_{{}_if} & = -K_f {}_if & \quad \text{if } h({}_i p) \neq 0. \IEEEyessubnumber \label{eq:force_control_dissipation}
\end{IEEEeqnarray}
When the point is in contact, $u_{{}_if}$ is free to take any value in $\left[-M_f, M_f\right]$ with $M_f \in \mathbb{R}^3$ defining control bounds (Eq. \eqref{eq:force_control_bounds}). On the other hand, if the contact point is not on the walking surface, the control input makes the contact force decreasing exponentially (Eq. \eqref{eq:force_control_dissipation}). Defining $\delta^*({}_ip)$ as a binary function such that
\begin{equation}
	\delta^*({}_ip) = 
	\begin{cases}
	1 & \quad \text{if } h({}_i p) = 0 \\
	0 & \quad h({}_i p) \neq 0
	\end{cases}, 
\end{equation}
it is possible to write Eq. \eqref{eq:force_control_cases} as a set of two inequalities:
\begin{IEEEeqnarray}{RCL}
	\IEEEyesnumber \phantomsection \label{eq:force_control_full}
- K_f \left(1 - \delta^*({}_ip)\right) {}_if - \delta^*({}_ip)M_f &\leq& u_{{}_if} \IEEEyessubnumber\\
- K_f \left(1 - \delta^*({}_ip)\right) {}_if + \delta^*({}_ip)M_f &\geq& u_{{}_if}. \IEEEyessubnumber
\end{IEEEeqnarray}
Even if $\delta^*({}_ip)$ would require the adoption of integer variables, it is possible to use a continuous approximation, $\delta({}_ip)$, namely the hyperbolic secant:
\begin{equation}\label{eq:sech}
    \delta({}_ip) = \text{sech}\left(\mathbf{k_h} h({}_i p)\right),
\end{equation}
where $\mathbf{k_h}$ is a user-defined scaling factor. Since $\delta({}_ip) = 0$ only when ${}_ip \rightarrow \infty$, Eq. \eqref{eq:force_control_full} satisfies the LICQ condition.

Given Eq. \eqref{eq:friction}, it is enough to apply Eq. \eqref{eq:force_control_full} only to the force component normal to the ground: if it decreases to zero, also planar force components have to vanish.

Since contact points are not supposed to penetrate the walking ground, we can impose $h({}_i p) \geq 0$.

None of the constraints defined above could prevent the contact points to move on the walking plane when in contact. In fact, even if friction constraints defined in Eq. \eqref{eq:friction} are satisfied, the contact points are still free to move on the contact surface. Force and position variables are (almost) independent at this stage. It is possible to prevent planar motions when in contact by limiting the effect of the control input $u_{{}_ip}$ along the planar components:
\begin{equation}\label{eq:planarControl}
	\mathbf{t}({}_ip)^\top {}_i \dot{p} = \tanh\left(\mathbf{k_t} h({}_ip)\right) \left[e_1 ~ e_2\right]^\top u_{{}_ip}
\end{equation}
where $\mathbf{k_t} \in \mathbb{R}$ is a user-defined scaling factor. Eq. \eqref{eq:planarControl} multiplies the control input along the planar direction to zero when $h({}_ip)$ is null and, at the same time, it will reduce the velocity when the contact point is approaching the ground. It is possible to rewrite Eq. \eqref{eq:planarControl} as
\begin{equation}
	{}_i\dot{p} =  \tau({}_ip) u_{{}_ip},
\end{equation}
where the function $\tau(\cdot): \mathbb{R}^3 \rightarrow \mathbb{R}^{3\times 3}$ is defined as:
\begin{equation}
    \tau({}_ip) ={}^\mathcal{I}R_{plane} ~ diag\left(\begin{bmatrix}
    \tanh\left(\mathbf{k_t} h({}i_p)\right) \\
     \tanh\left(\mathbf{k_t} h({}i_p)\right) \\
      1
    \end{bmatrix}\right).
\end{equation}
Note that, from now on, $u_{{}_ip}$ is assumed to be defined in $plane$ coordinates. Thus, the normal component of the velocity is directly affected by $e_3^\top u_{{}_ip}$. Also, it is necessary to bound this control input, $u_{{}_ip} \in \left[-M_V, M_V\right], M_V \in \mathbb{R}^3$, to properly exploit the effect of the hyperbolic tangent. Note that Eq. \eqref{eq:planarControl} allows avoiding the use of complementarity conditions along planar directions.

\subsection{Contact Point Position Consistency}

While each contact point is supposed to be independent from the control point of view, they all need to remain on the same surface and maintain a constant relative distance, since they belong to the same rigid body. At the same time, we want them to be within the workspace reachable by the robot legs. We can achieve both the objectives by enforcing the following constraint on each of the contact points:
\begin{equation}
	{}_ip = {}^\mathcal{I} H_\text{foot} {}^\text{foot}{}_i p,
\end{equation} 
where ${}^\text{foot}{}_i p$ is the (fixed) position of the contact point within the foot surface, expressed in foot coordinates. Here, the transformation matrix ${}^\mathcal{I} H_\text{foot}$ would depend on the base position ${}^\mathcal{I}p_B \in \mathbb{R}^3$, the base quaternion ${}^\mathcal{I}\rho_B \in \mathbb{H}$ and the joints configuration $s \in \mathbb{R}^n$, with $n$ being the number of joints. As a consequence, the full kinematics of the robot is taken into consideration and the following dynamic equations have to be considered:
\begin{IEEEeqnarray}{RCL}
	\IEEEyesnumber \phantomsection
	{}^\mathcal{I}\dot{p}_b &=& {}^\mathcal{I}R_B {}^Bv_{\mathcal{I},B} \IEEEyessubnumber \label{eq:quaternionLeftDer}\\
	{}^\mathcal{I}\dot{\rho}_B &=& u_\rho \IEEEyessubnumber \label{eq:quaternionRotationDerivative} \\
	\dot{s} &=& u_s \IEEEyessubnumber.
\end{IEEEeqnarray}
Here ${}^Bv_{\mathcal{I},B} \in \mathbb{R}^3$, $u_\rho \in \mathbb{R}^4$ and $u_s \in \mathbb{R}^n$ are considered control inputs defining the base linear velocity, the quaternion derivative and the joints velocity, respectively. More specifically, ${}^Bv_{\mathcal{I},B}$ is the linear part of ${}^BV_{\mathcal{I},B} \in \mathbb{R}^6$ the left-trivialized (i.e. measured in body coordinates) base velocity.

\subsection{Momentum Dynamics}
In Sec. \ref{sec:contacts_interface}, we consider the contact points as if they have the possibility of exerting a force with the environment. We can describe the effect of these contact forces through the momentum, or centroidal, dynamics. This choice is supported by the fact that the momenutm dynamics depends only on the contact forces, their location and on the center of mass (CoM) position, $x_\text{CoM} \in \mathbb{R}^3$. Define ${}_{\bar{G}} h \in \mathbb{R}^6$ as the robot total momentum, with
${}_{\bar{G}} h = \left[
{}_{\bar{G}} h^{p\top} ~
{}_{\bar{G}} h^{\omega\top}
\right]^\top$
where ${}_{\bar{G}} h^p \in \mathbb{R}^3$ and ${}_{\bar{G}} h^\omega \in \mathbb{R}^3$ are respectively the linear and angular momentum. This quantity is expressed in a frame oriented as the inertial frame $\mathcal{I}$, with the origin placed on the CoM position. Such frame is called $G[\mathcal{I}]$ or simply $\bar{G}$. The momentum dynamics has the following form:
\begin{IEEEeqnarray}{RCL}
	\IEEEyesnumber \phantomsection
	{}_{\bar{G}} \dot{h} &=& m\bar{g} + \sum_i
	\begin{bmatrix}
		\mathds{1}_3 \\
		S({}_ip - x_{\text{CoM}})
	\end{bmatrix} {}_if \IEEEyessubnumber\\	
	\dot{x}_\text{CoM} &=& \frac{1}{m}  \left({}_{\bar{G}} h^p\right)  \IEEEyessubnumber \label{eq:CoMDynamics}
\end{IEEEeqnarray}
with $m$ the total mass of the robot, $\bar{g} = \left[\begin{smallmatrix} 0 & 0 & -g & 0 & 0 & 0\end{smallmatrix}\right]^T$. We also need to make sure that the integrated CoM corresponds to the one obtained via the joint variables. This is done through an additional constraint:
\begin{equation}
x_\text{CoM} = \text{CoM}({}^\mathcal{I}p_B, {}^\mathcal{I}\rho_B, s)
\end{equation} 
where $\text{CoM}({}^\mathcal{I}p_B, {}^\mathcal{I}\rho_B, s)$ is the function mapping base pose and joint positions to the CoM position. While this constraint defines a link between the linear momentum and the joint variables, the same would not hold for the angular part. To this end, we can exploit the Centroidal Momentum Matrix \cite{orin08} ($J_\text{CMM}$). In fact, the robot angular momentum can be defined as a function of the base and joints velocity:
\begin{equation}
{}_{\bar{G}} h^\omega = \left[0_{3 \times 3} ~ \mathds{1}_3\right]J_\text{CMM} \nu
\end{equation}
where $ \nu = \left[{}^BV_{\mathcal{I},B} ~ u_s\right]^\top$. Here, the base angular velocity ${}^B \omega_{\mathcal{I},B}$ can be substituted with the quaternion derivative through the map $\mathcal{G}$ \cite[Section 1.5.4]{graf2008quaternions}, such that 
\begin{equation*}
	{}^B \omega_{\mathcal{I},B} = 2 \mathcal{G}({}^\mathcal{I}\rho_B)u_\rho.
\end{equation*}

Some additional constraints can be considered:
\begin{IEEEeqnarray}{RCL}
    \IEEEyesnumber
    	x_{\text{CoM},z \text{ min}} &\leq& e_3^\top{x_{\text{CoM}}}  \IEEEyessubnumber \label{constr:com_height_limit}\\
	-M_{h_\omega} &\leq& {}_{\bar{G}} h^\omega \leq M_{h_\omega} \IEEEyessubnumber \label{constr:angular_momentum_bounds}
\end{IEEEeqnarray}
Eq \eqref{constr:com_height_limit} avoids solutions which would bring the CoM position too close or below the ground. Eq. \eqref{constr:angular_momentum_bounds} provides an upper and lower bound $M_{h_\omega} \in \mathbb{R}^3$ to the angular momentum. These constraints avoid trajectories that would cause excessive motions or let the robot falling.

\subsection{Feet Minimum Lateral Distance}
While taking steps, we need to make sure that the robot legs do not collide with each other. Self collision constraints are usually hard to be considered and may slow down consistently the determination of a solution. A simpler solution consists in avoiding the left leg to be on the right of the other leg. Consequently, cross steps are forbidden. Let us consider a frame attached to the right foot with the positive $y-$direction pointing toward left. In this case, it is sufficient to impose the $y-$component of the ${}^{r}x_{l}$ (i.e. the relative position of the left foot expressed in the right foot frame) to be greater than a given quantity, i.e. $e_2^\top {}^{r}x_{l} \geq d_\text{min}$.

%% file: tex/Tasks.tex
\section{Tasks} \label{sec:tasks}
We present the set of tasks used to plan a walking trajectory. While constraints define the model and the control limitations, the tasks embed the planning objectives.

\subsection{Contact point centroid position task}
In order to make the robot moving toward a desired position, we minimize the L2 norm of the error between a point attached to the robot and its desired position in an absolute frame. 
In particular, we select the centroid of the contact points as target, thus avoiding to specify a desired placement for each foot:
\begin{equation}
\Gamma_{{}_\# p} = \frac{1}{2} \|{}_\# p - {}_\# p^*\|^2_{W_\#}
\end{equation}
where $\#$ is the number of contact points in a single foot. Thus, we have ${}_\# p =\frac{1}{2\#} \sum_{l, r}\sum^{\#}_i {}_ip$ and ${}_\# p^* \in \mathbb{R}^3$ is a user-defined reference value.

\subsection{CoM linear velocity task}
While walking, we want the robot to keep a constant forward motion. In fact, since the positions of the feet are not scripted, it may be possible to plan two consecutive steps with the same foot. By requiring a constant forward velocity, such phenomena can be avoided. This task is defined as:
\begin{equation}
\Gamma_{{}_{\bar{G}} h^p} = \frac{1}{2} \|{}_{\bar{G}} h^p - m  v_G^*\|^2_{W_v}
\end{equation}
with $v_G^* \in \mathbb{R}^3$ a desired CoM velocity. The matrix $W_v$ selects and weights the different directions separately.

\subsection{Frame orientation task}
While moving, we want a robot frame to be oriented in a specific orientation ${}^\mathcal{I}R^*_\text{frame}$. In particular, we weight the distance of the rotation matrix ${}^\mathcal{I}\tilde{R}_\text{frame} = {}^\mathcal{I}R^{*\top}_\text{frame}{}^\mathcal{I}R_\text{frame}$ from the identity. Having to express this task in vector form, we convert ${}^\mathcal{I}\tilde{R}_\text{frame}$ into a quaternion (through the function \texttt{quat} which implements the Rodriguez formula) and weight its difference from the identity quaternion $I_q$. Namely:
\begin{equation}\label{cost:frameOrientation}
\Gamma_\text{frame} = \frac{1}{2}\left\|\texttt{quat}\left({}^A\tilde{R}_\text{frame}\right) - I_q\right\|^2.
\end{equation}
This corresponds to a simplified version of the quaternion difference metric listed in \cite{huynh2009metrics}, under the assumption for $\texttt{quat}$ to always return a quaternion with positive real value.

\subsection{Force regularization task} \label{sec:forceRegularization}
While considering each single contact force in a foot as independent, they still belong to a single body part. Thus, we prescribe the contact forces in a foot to be as similar as possible, refraining from using partial contacts if not strictly necessary. This can be obtained through the following:
\begin{equation}\label{cost:forceRegularization}
\Gamma_{\text{reg} f} = \sum_{l, r}\sum^{\#}_i \frac{1}{2} \left\|{}_if - \frac{1}{\#}\sum^{\#}_j {}_jf\right\|^2.
\end{equation}

\subsection{Joint regularization task}
The joint configuration $s$ is part of the optimization variables. In order to prevent the planner from providing solutions with huge joint variations, we introduce a regularization task for joint variables:
\begin{equation}\label{cost:jointsRegularization}
\Gamma_{\text{reg} s} = \frac{1}{2}\left\|\dot{s} + K_s(s - s^*)\right\|^2_{W_j}
\end{equation}
with $s^*$ the desired joint configurations and $W_j$ a weight matrix. The minimum of this cost is when $\dot{s}= -K_s(s - s^*)$, with $K_s \in \mathbb{R}^{n\times n}$. When this equality holds, joint values converge exponentially to their reference $s^*$. Hence, joint velocities and joint positions are regularized at the same time.

\subsection{Swing height task}
When performing a step, the swing foot clearance usually ensures some level of robustness with respect to ground asperity. Nevertheless, since the soil profile is supposed to be known in advance, a solution satisfying all the equations described in Sec. \ref{sec:modeling} may require the swing foot to be raised just few millimeters from the ground. In order to specify a desired swing height, we impose the following cost:
\begin{equation}
	\Gamma_{\text{swing}} = \sum_{l, r}\sum^{\#}_i\frac{1}{2}\left\|\left(e_3^\top {}_ip - {}_sh^*\right)\left[e_1 ~ e_2\right]^\top u_{{}_ip}\right\|.
\end{equation}
It penalizes the distance between the $z$-component of each contact point position from a desired height ${}_sh^* \in \mathbb{R}$ when the corresponding planar velocity is not null. Trivially, this cost has two minima: when the planar velocity is zero (thus the point is not moving) or when the height of the point is equal to the desired one.

%% file: tex/OptimalControlProblem.tex
\section{Optimal Control Problem} \label{sec:oc}
Given the set of equations listed in Sec. \ref{sec:modeling} and the tasks described in Sec. \ref{sec:tasks} it is possible to define an optimal control problem, whose complete formulation is presented below. Here, the vector $\mathbf{w}$ contains the set of weights defining the relative ``importance'' of each task.
\subsection{Problem definition} \label{sec:problem_definition}
\begin{IEEEeqnarray*}{CC}
	\nonumber
	\minimize_{\mathcal{X}, \mathcal{U}} & 
	\left[
		\Gamma_{{}_\# p} ~
		\Gamma_{{}_{\bar{G}} h^p} ~
		\Gamma_\text{frame} ~
		\Gamma_{\text{reg} f} ~
		\Gamma_{\text{reg} s} ~
		\Gamma_{\text{swing}} 
	\right]^\top  \mathbf{w} \label{costFunction}\\
	\text{subject to : }&\hspace*{0.7\columnwidth} \nonumber
\end{IEEEeqnarray*}
$\bullet$ Dynamical Constraints 
\begin{IEEEeqnarray}{RCLL}
	\IEEEyesnumber \phantomsection \label{constr:system_dynamics}
	{}_i\dot{f} &=& u_{{}_if} & \forall \text{ contact point} \IEEEyessubnumber \label{constr:force_derivative}\\
	{}_i\dot{p} &=& \tau({}_ip) u_{{}_ip} & \forall \text{ contact point}  \IEEEyessubnumber \label{constr:position_derivative}\\
	{}_{\bar{G}} \dot{h} &=& \IEEEeqnarraymulticol{2}{L}{m\bar{g} + \sum_i
		\begin{bmatrix}
			\mathds{1}_3 \\
			S({}_ip - x_{\text{CoM}})
		\end{bmatrix} {{}_if}} \IEEEyessubnumber\label{constr:momentum_derivative}\\	
	\dot{x}_\text{CoM} &=& \IEEEeqnarraymulticol{2}{L}{\frac{1}{m}  \left({}_{\bar{G}} h^p\right)}  \IEEEyessubnumber \label{constr:com_derivative}\\
	{}^\mathcal{I}\dot{p}_B &=& {}^\mathcal{I}R_B {}^Bv_{\mathcal{I},B} \IEEEyessubnumber \label{constr:basePosDerivative}\\
	{}^\mathcal{I}\dot{\rho}_B &=& u_\rho \IEEEyessubnumber \label{constr:quatDerivative}\\
	\dot{s} &=& u_s \IEEEyessubnumber
\end{IEEEeqnarray}
$\bullet$ Equality Constraints 
\begin{IEEEeqnarray}{RCL}
	\IEEEyesnumber \phantomsection
	{}_ip &=& {}^A H_\text{foot} {}^\text{foot}{}_i p \quad \forall \text{ contact point} \IEEEyessubnumber \label{constr:pointPositions}\\
	x_\text{CoM} &=& \text{CoM}({}^\mathcal{I}p_B, {}^\mathcal{I}\rho_B, s) \IEEEyessubnumber \\
	{}_{\bar{G}} h^\omega &=& \left[0_{3 \times 3} ~ \mathds{1}_3\right]J_\text{CMM} \begin{bmatrix}
		{}^Bv_{\mathcal{I},B} \\
		2 \mathcal{G}({}^\mathcal{I}\rho_B)u_\rho\\
		u_s
	\end{bmatrix} \IEEEyessubnumber \label{constr:CMM} \\
	\|{}^\mathcal{I}\rho_B\|^2 &=& 1 \IEEEyessubnumber 
\end{IEEEeqnarray}
$\bullet$ Inequality Constraints
\begin{IEEEeqnarray}{RCL}
	\IEEEyesnumber \phantomsection \label{constr:inequalities}
	\mathbf{n}({}_ip)^\top{}_if &\geq& 0 \IEEEyessubnumber\\
	\|\mathbf{t}({}_ip)^\top {}_if \| &\leq& \rho ~\mathbf{n}({}_ip)^\top{}_if \IEEEyessubnumber\\
   u_{{}_if} &\geq& - K_f \left(1 - \delta({}_ip)\right) {}_if - \delta({}_ip)M_f   \IEEEyessubnumber\\
   u_{{}_if} &\leq&  - K_f \left(1 - \delta({}_ip)\right) {}_if + \delta({}_ip)M_f  \IEEEyessubnumber \\
	-M_V &\leq& u_{{}_ip} \leq M_V \IEEEyessubnumber\\
	h({}_i p) &\geq& 0 \IEEEyessubnumber\\
	e_2^\top{{}^{r}x_{l}} &\geq& d_\text{min} \IEEEyessubnumber \label{constr:minDistance}\\
	x_{\text{CoM},z \text{ min}} &\leq& e_3^\top{x_{\text{CoM}}}  \IEEEyessubnumber\\
	-M_{h_\omega} &\leq& {}_{\bar{G}} h^\omega \leq M_{h_\omega} \IEEEyessubnumber
\end{IEEEeqnarray}
Here, the state variables $\mathcal{X}$ are those derived in time, $\mathcal{U}$ all the others. More specifically:
\begin{equation}
\mathcal{X} = 
\begin{bmatrix}
{}_if \\
{}_ip \\
\vdots	\\
{}_{\bar{G}} h \\
x_\text{CoM} \\
{}^\mathcal{I} p_B \\
{}^\mathcal{I}\rho_B \\
s
\end{bmatrix}, \quad
\mathcal{U} = 
\begin{bmatrix}
u_{{}_if} \\
u_{{}_ip} \\
\vdots \\
{}^Bv_{\mathcal{I},B} \\
u_\rho \\
u_s
\end{bmatrix}
\end{equation}
where the symbol $\vdots$ represents the repetition of variables for each contact point. 
The optimal control problem is solved using a Direct Multiple Shooting method \cite{betts2010practical}. The system dynamics, defined in Eq. \eqref{constr:system_dynamics}, is discretized adopting an implicit trapezoidal method with a fixed integration step. The corresponding optimization problem is solved thanks to \texttt{Ipopt} \cite{wachter2006implementation}. 

The walking trajectories are generated using the Receding Horizon Principle \cite{mayne1990receding}, adopting a fixed prediction window.

\subsection{Considerations} \label{sec:considerations}
The optimal control problem described in Sec. \ref{sec:problem_definition} is built such that (almost) no constraint is task specific. As a consequence, it is particularly important to define the cost function carefully since the solution will be a trade-off between all the various tasks. On the other hand, the detailed model of the system allows achieving walking motions without specifying a desired CoM trajectory or by fixing the angular momentum to zero.
Nevertheless, due to the limited time horizon, it is better to prevent the solver from finding solutions which would bring the robot to unfeasible states in future planner iterations. To this end, Eq. \eqref{constr:com_height_limit} and Eq. \eqref{constr:angular_momentum_bounds} have been added, using reasonably large bounds. 

Another possible effect resulting from the application of the Receding Horizon principle, is the emergence of ``procrastination'' phenomena. Due to the moving horizon, the solver may continuously delay in actuating motions, since the task keeps being shifted in time. A simple fix to this phenomena is to increase the weights $\mathbf{w}$ with time, such that it is more convenient to reach a goal position earlier.

Finally, given that the problem under consideration is non-convex, the optimizer will find a local minimum. This may result in a sub-optimal solution for the given tasks. 

During the first iteration, the solver is initialized by simply translating the whole robot in the desired position. In successive iterations, the solver is warm-started with the solution previously computed.

%% file: tex/Experiments.tex
\section{Results} \label{sec:experiments}

\begin{figure}[tpb]
    \centering
    \subfloat[$t=0.5s$] {\includegraphics[width=.22\columnwidth]{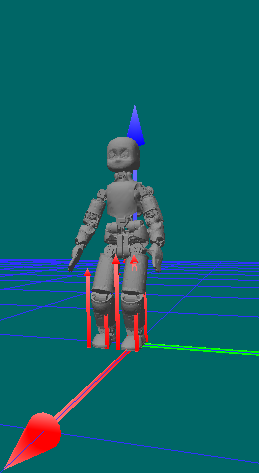}}
    \hspace{.001\columnwidth}
    \subfloat[$t=1.5s$] {\includegraphics[width=.22\columnwidth]{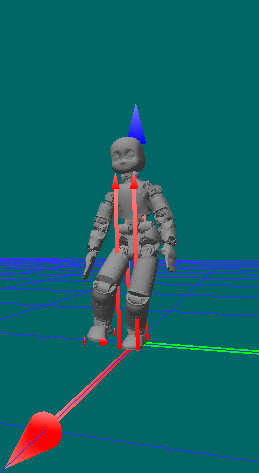}}
    \hspace{.001\columnwidth}
    \subfloat[$t=2.5s$] {\includegraphics[width=.22\columnwidth]{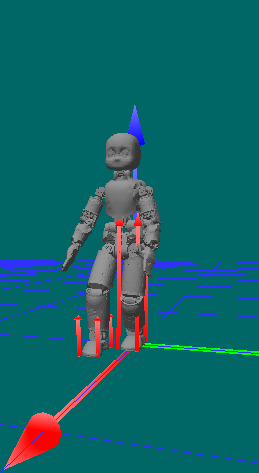}}
    \hspace{.001\columnwidth}
    \subfloat[$t=3.5s$] {\includegraphics[width=.22\columnwidth]{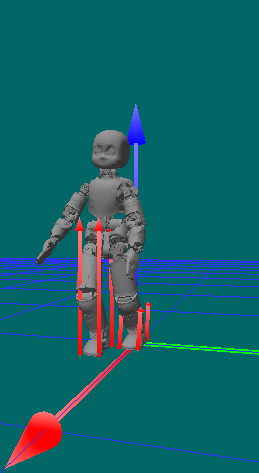}}
    \caption{Snapshots of the generated walking motion. The red arrows indicate the force required at each contact point scaled by a factor of 0.01. }
    \label{fig:slow_straight}
\end{figure}

\begin{figure}[tpb]
    \centering
    \includegraphics[width=0.95\columnwidth]{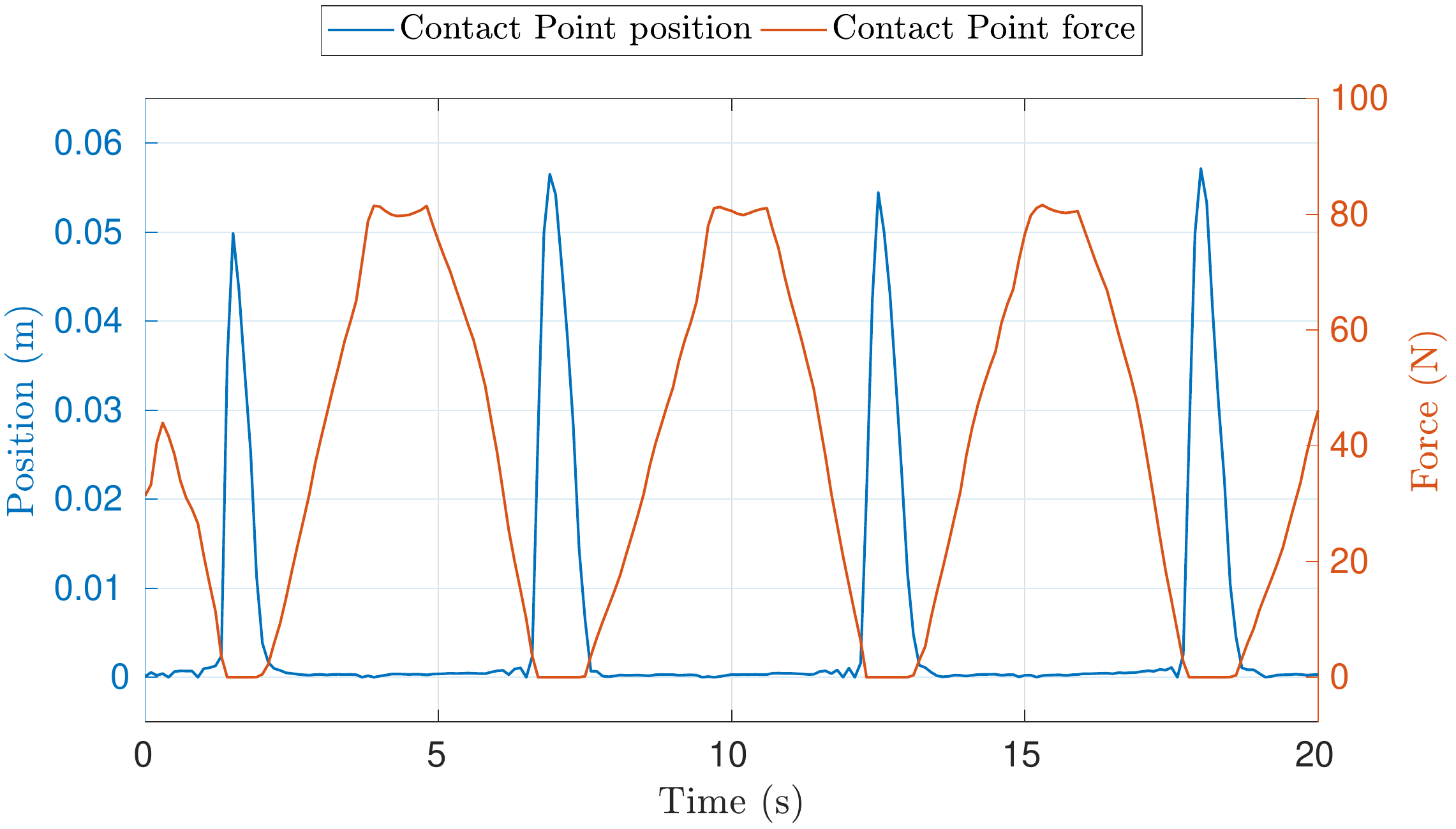}
    \caption{Normal force and normal position of a contact point of the right foot plotted together.}
    \label{fig:point_positionVsForce}
\end{figure}

\begin{figure}[tpb]
    \centering
    \includegraphics[width=0.95\columnwidth]{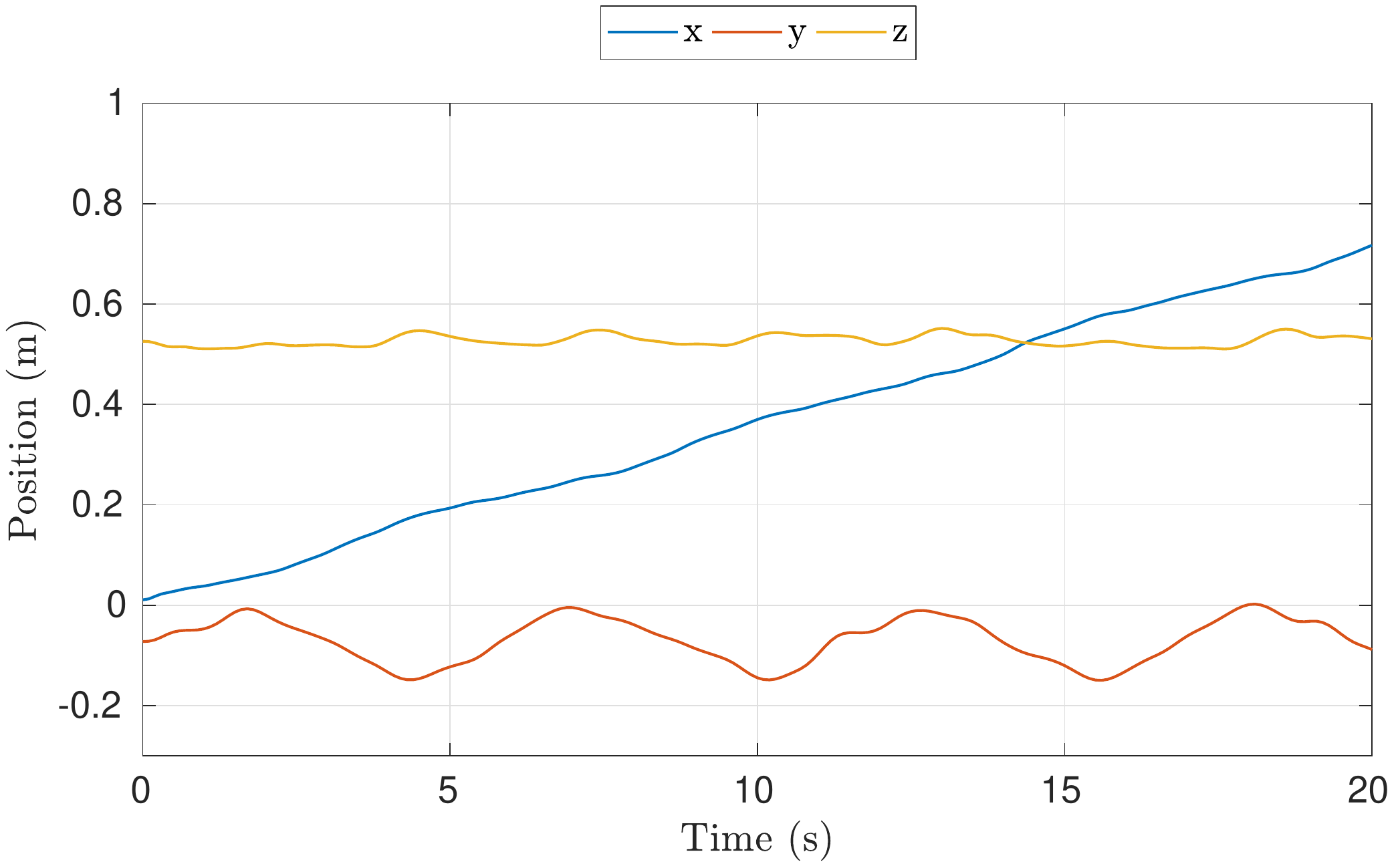}
    \caption{Planned CoM position.}
    \label{fig:com_position}
\end{figure}

\begin{figure}[tpb]
    \centering
    \includegraphics[width=0.95\columnwidth]{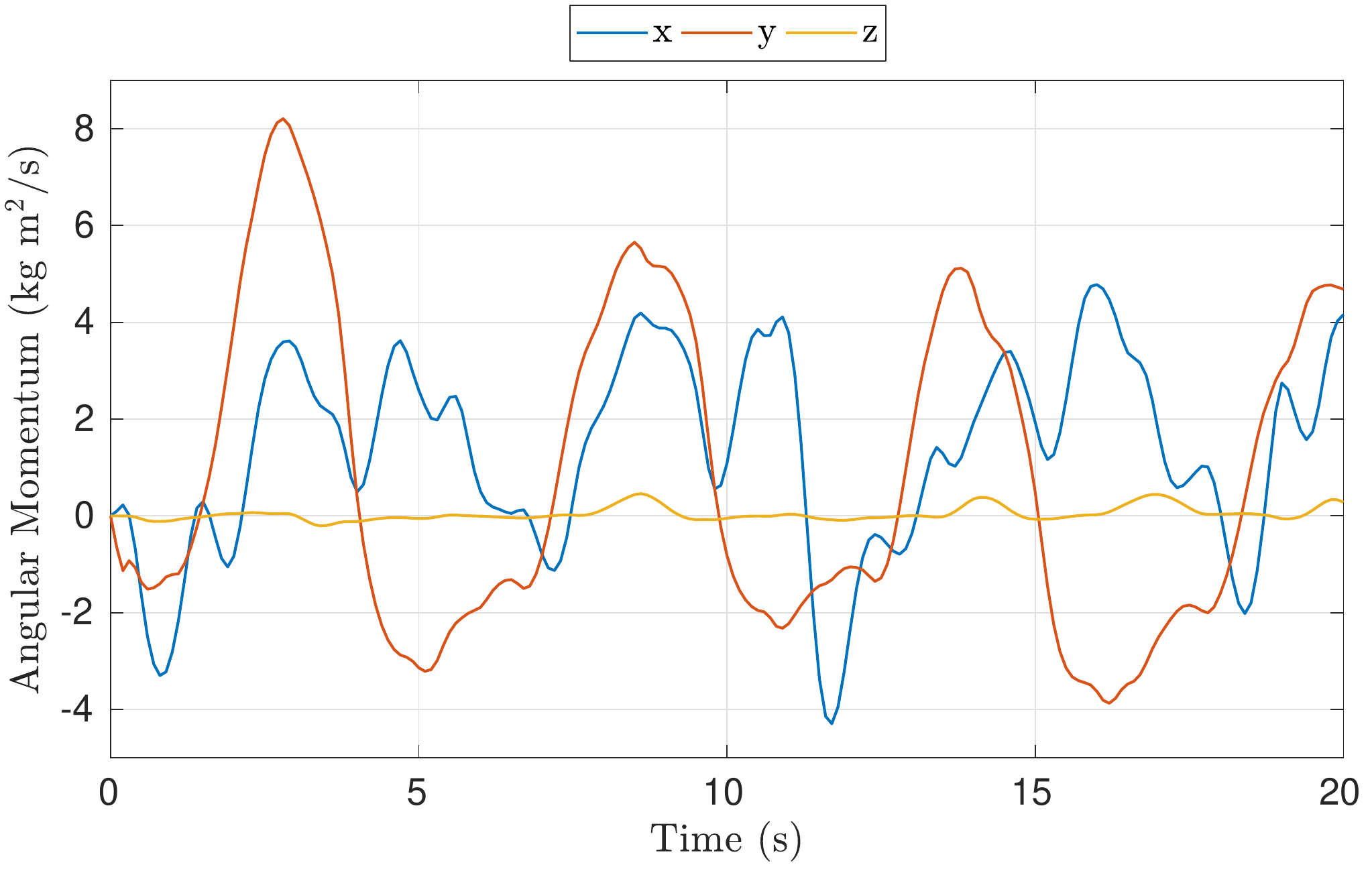}
    \caption{Planned angular momentum.}
    \label{fig:angular_momentum}
\end{figure}

The optimal control problem described in Sec. \ref{sec:oc} is used to generate a walking motion. The integration step is set to $100 \texttt{ms}$, while the prediction horizon is $2\texttt{s}$. After each iteration of the MPC controller, the previously computed state is used as a feedback. Regarding the scaling factors, we use $\mathbf{k_h} = 300$ and $\mathbf{k_t} = 10.0$. They appeared to be reasonable values for having meaningful simulations of contacts. In addition, these values are robot and ground independent, since they depend only on the position of a contact point with respect to the ground.

The trajectories have been generated using the iCub humanoid \cite{Nataleeaaq1026} robot model on a 7$^{th}$ generation Intel\textsuperscript{\textregistered} Core i7@2.8GHz laptop.
We assume the ground to be flat, while we control 23 of the robot joints. For each foot, we consider four contact points located at the vertexes of the rectangle enclosing the robot foot. Concerning the references, the desired position for the centroid of the contact points is moved $10 \texttt{cm}$ along the walking direction every time the robot performs a step. A simple state machine, where the reference is moved as soon as a step is completed, is enough to generate a continuous walking pattern. The speed is modulated by prescribing a fixed desired CoM forward velocity equal to $5 \texttt{cm/s}$.

Figure \ref{fig:slow_straight} shows some snapshots of the first generated step. Also, it can be observed the effect of the contact parametrization described in Sec. \ref{sec:contact_parametrization} from Fig. \ref{fig:point_positionVsForce}. The normal force decreases to zero as soon as the foot starts leaving the ground, and then it grows again at touchdown. It is possible to recognize the different walking phases, even though they are not planned a priori.

Figure \ref{fig:com_position} presents the planned CoM position. Here, it is possible to notice that $x$ position grows at a constant rate. This is a direct consequence of the task on the CoM velocity. 
Figure \ref{fig:angular_momentum} shows the planned angular momentum, which is not fixed to zero. Although it is limited to $10~\texttt{kg}~{\texttt{m}^2}/{\texttt{s}}$, such limit is never reached. Similarly, the bound on the CoM height, $x_{\text{CoM},z \text{ min}}$, is set to half of the initial robot height, but such constraint is never activated.

It is worth stressing that none of the tasks described above define how and when to raise the foot. By prescribing a reference for the centroid of the contact points and by preventing the motion on the contact surface, swing motions are planned automatically. Nevertheless, this advantage comes with a cost. It is difficult to define a desired swing time and, more importantly, the relative importance of each task, i.e. the values of $\mathbf{w}$, must be chosen carefully. During experiments, we adopted an incremental approach. We added the tasks one by one, starting from $\Gamma_{{}_\# p}$ and then we gradually refine the walking motion by tuning a cost at a time.

%% file: tex/Conclusions.tex
\section{Conclusions} \label{sec:conclusions}
This paper presents a planner capable of generating walking trajectories using a minimal set of references. It considers the centroidal momentum of the robot and its full kinematics to plan dynamically consistent step motions. The modelling of contacts makes use of a novel parametrization approach, allowing to model the interface between the robot and the ground with a set of continuous equations. 
Currently, this model does not consider impacts nor contact sliding. The consideration of slip-turn motions \cite{miura2013slip} is left as future work.

The results have to be considered as an initial validation of the generated trajectories when the iCub model is adopted.
In particular, it is shown that walking trajectories can emerge by specifying a moving reference for the centroid of the contact points and the desired CoM velocity only.

The planner considers relatively large time-steps. This enables the insertion of another control loop at higher frequency, whose goal is to stabilize the planned trajectories. As a future work, we will consider connecting this planner to the whole-body controller presented in \cite{dafarra2018control}.

The main bottleneck is represented by the computational time. A single planner iteration may take from slightly less than a second to more than a minute. This prevents an online implementation on the real robot. On the other hand, the continuous formulation of the optimal control problem allows the application of techniques, like those presented in \cite{neunert2018whole, farshidian2017real}, which do not rely on the transcription to a non-linear programming problem.